\documentclass[10pt,twocolumn,letterpaper]{article}

\usepackage[pagenumbers]{cvpr} % To force page numbers, e.g. for an arXiv version

\definecolor{cvprblue}{rgb}{0.21,0.49,0.74}
\usepackage[pagebackref,breaklinks,colorlinks,allcolors=cvprblue]{hyperref}

\usepackage{amsmath}
\usepackage{array}
\usepackage{booktabs}
\usepackage{multirow}
\usepackage{amsmath}  % for advanced math typesetting
\usepackage{amsfonts} % for \mathbb and other fonts
\usepackage{amssymb}  % for additional symbols (optional)
\usepackage{algpseudocode} % or algorithmic
\usepackage{algorithm}
\usepackage{adjustbox}
\usepackage{xcolor}
\usepackage[normalem]{ulem}
\def\paperID{13588} % *** Enter the Paper ID here
\def\confName{CVPR}
\def\confYear{2025}

\title{Secure Generalization through Stochastic Bidirectional Parameter Updates Using Dual-Gradient Mechanism}

\author{Shourya Goel, Himanshi Tibrewal, Anant Jain, Anshul Pundhir, Pravendra Singh\\
Department of Computer Science and Engineering\\
Indian Institute of Technology Roorkee, Uttarakhand, India\\
{\tt\small \{shourya\_g, anant\_j, himanshi\_t, anshul\_p, pravendra.singh\}@cs.iitr.ac.in}
}

\begin{document}
\maketitle
\begin{abstract}
Federated learning (FL) has gained increasing attention due to privacy-preserving collaborative training on decentralized clients, mitigating the need to upload sensitive data to a central server directly. Nonetheless, recent research has underscored the risk of exposing private data to adversaries, even within FL frameworks. In general, existing methods sacrifice performance while ensuring resistance to privacy leakage in FL. We overcome these issues and generate diverse models at a global server through the proposed stochastic bidirectional parameter update mechanism. Using diverse models, we improved the generalization and feature representation in the FL setup, which also helped to improve the robustness of the model against privacy leakage without hurting the model's utility. We use global models from past FL rounds to follow systematic perturbation in parameter space at the server to ensure model generalization and resistance against privacy attacks. We generate diverse models (in close neighborhoods) for each client by using systematic perturbations in model parameters at a fine-grained level (i.e., altering each convolutional filter across the layers of the model) to improve the generalization and security perspective. We evaluated our proposed approach on four benchmark datasets to validate its superiority. We surpassed the state-of-the-art methods in terms of model utility and robustness towards privacy leakage. We have proven the effectiveness of our method by evaluating performance using several quantitative and qualitative results. 
\end{abstract}    
\section{Introduction}\label{sec:introduction}

In recent years, Federated Learning (FL)~\cite{mcmahan2017communication} has gained wide attention across various domains, including healthcare~\cite{guo2021multi,liu2021feddg,jiang2022harmofl}, autonomous driving~\cite{xie2022efficient}, etc., since FL allow clients to locally train data and share only model parameters (not sensitive data) to the global server for aggregation. Current studies~\cite{geng2021towards,fu2022label,yu2023untargeted,zhu2023surrogate} highlighted the issue of privacy leakage through shared model parameters, which offers vulnerability to adversaries in the form of different types of attacks. Several attempts have been made to solve the privacy-leakage issues and provide enhanced protection, which includes homomorphic encryption~\cite{aono2017privacy,jin2023fedml}, differential privacy~\cite{abadi2016deep,liao2023ppgencdr}, and gradient perturbation~\cite{sun2021soteria}. These methods attempt to secure the privacy of sensitive data at the cost of computational overhead or sacrifice the model's efficiency. Researchers aim to maintain utility without sacrificing the model's accuracy and encrypt the training data~\cite{chuman2018encryption,huang2020instahide}.
% To maintain utility without sacrificing the model's accuracy, researchers focus on encrypting the training data~\cite{chuman2018encryption,huang2020instahide}. 
These approaches require sharing classifier model parameters to perform model aggregation at the server and defend against image reconstruction attacks in FL. Following these attempts, Researchers have proven the vulnerability of clients in these methods for label inference attacks and membership inference attacks, and hence not suitable to provide adequate security~\cite{shokri2017membership,geng2021towards}.

Recently, Yutin et al.~\cite{ma2024ppidsg} provided a theoretical analysis of different attacks in FL and highlighted the concerns of privacy leakage due to the sharing of classifier parameters. To overcome the privacy leakage issue, Yutin et al.~\cite{ma2024ppidsg} proposed a Generative Adversarial Network (GAN) based privacy-preserving image distribution sharing scheme (PPIDSG) in FL, which does not require the sharing of classifier model parameters. To secure federated learning, PPIDSG employs GAN-based parameter sharing to learn the distribution of encrypted images and update client models with an aggregated model. However, learning in the encrypted domain involves a trade-off between utility and security. Also, the same global update to different clients limits the generalization of clients. Moreover, the gradients communicated from the global server to clients are also susceptible to various attacks, which can be improved to make the model more secure. Particularly, to avoid privacy leakage, researchers have proposed differential privacy (DP)~\cite{abadi2016deep,liao2023ppgencdr,wei2020federated} and gradient perturbation-based methods~\cite{zhu2019deep,sun2021soteria}. The aforementioned approaches ensure resistance to privacy leakage but also sacrifice the performance of FL due to a lack of systematic perturbation in gradients.

Motivated by these observations and considering these gaps, we found scope to improve the utility and security perspectives in FL. Particularly, we focus on: 1) How to retain the model utility when focusing on security by not sharing classifier model parameters in FL communications. 2) How to improve the robustness of the FL setup against different attacks without sacrificing the utility of the model in terms of its classification accuracy. To achieve these objectives, we proposed a novel approach that provides a more generalized and robust update from the global model to clients during FL, which improves the robustness of the model against the different attacks and does not sacrifice the model's utility while making the model secure. Our stochastic bidirectional update approach uses a dual-gradient mechanism to generate diverse models (in close neighborhoods) for each client, which improves the generalization and security perspective of FL. After obtaining diverse global models, we do not make any further alterations that help retain utility and generalization.

\noindent
\textbf{Our Contributions:} We propose a novel approach that follows our stochastic bidirectional parameter update mechanism to generate diverse and generalizable global models for different clients. The proposed approach improves the robustness of clients in FL against different data attacks without sacrificing the model's utility. Our approach makes systematic alterations to the global model using a dual gradient mechanism to make multiple diverse models by using global models from previous FL rounds. The diverse models generated by our method are in a close neighborhood so that clients can improve generalization as well as robustness against privacy attacks. We validated the superiority of our approach using four datasets against state-of-the-art (SOTA) methods. Our method is evaluated considering the model's utility and robustness against attacks and surpasses the SOTA methods.

\section{Related work}\label{sec:related_work}
Several optimization approaches have been proposed to improve the utility of FL methods. The various optimization methods for FL can be categorized into global variable-based~\cite{li2020federated,karimireddy2020scaffold}, device grouping-based~\cite{fraboni2021clustered,chen2020fedcluster,li2021privacy} knowledge distillation-based~\cite{sattler2021fedaux,lin2020ensemble,zhu2021data}. In FedProx~\cite{li2020federated}, a proximal term is calculated as a squared distance of a global model with local models that helps in regularizing local loss and helps in model convergence. SCAF-FOLD~\cite{karimireddy2020scaffold} improves local training through global control variables to adjust optimization direction in each round of FL. The device-grouping FL approaches optimize local training by heuristic-based selection of local devices from the device groups for local training, which are grouped based on the specific similarity metric (model similarity). CluSamp~\cite{fraboni2021clustered} performs the client grouping based on sample size or model similarity. FedCluster~\cite{chen2020fedcluster} follows cyclic FL, wherein in each FL round, clients are grouped into multiple groups that perform FL. 

Knowledge distillation-based methods help to improve the inference of the FL by using knowledge of the teacher network to teach the student network. FedAUX~\cite{sattler2021fedaux} makes use of an auxiliary dataset for knowledge distillation and initialize server model. FedDF~\cite{lin2020ensemble} accelerates the FL by using the ensemble model as a teacher model and unlabelled data for knowledge distillation. The global variable-based methods are computationally demanding in terms of additional communication of global variables and proximal term computation over clients. The device grouping methods need to access all local methods to estimate similarity for grouping, which leads to vulnerability for privacy leakage. On the other hand, the knowledge distillation-based methods need additional overhead for computations and datasets for the distillation process.

To ensure security in FL, researchers have proposed various types of defense mechanisms against the attacks. The common attacks in FL include property inference attacks, membership inference attacks, and image reconstruction attacks. In the property inference attack, the adversary aims to determine the specific attributes that belong to a subset of the training data~\cite{fu2022label}. In a label inference attack, the adversary aims to determine the label attribute. In an image reconstruction attack, the adversary uses gradients sent from the client to the server model to reconstruct the original image. The authors attempted to perform minimization optimization using gradient difference for the original image and dummy image DLG~\cite{zhu2019deep}, which was further enhanced through the extraction of ground truth labels in iDLG~\cite{zhao2020idlg}. Gradient Inversion~\cite{yin2021see} is proposed to reconstruct complex and high-fidelity images using group consistency regularization. 

To make secure FL, researchers have also utilized GAN-based methods. The Generative Adversarial Network (GAN) was proposed to generate images resembling real ones using min-max optimization and adversarial loss between the generator and discriminator network~\cite{goodfellow2014generative}. GAN can also be used for image translation, which has also been explored by researchers for defense or attack. The adversary can use GAN to generate the target distribution images in real time~\cite{hitaj2017deep}. To perform a model extraction attack through a substitute network trained using GAN~\cite{zhao2020idlg}. Conditional GANs were utilized in FedCG to resist the image reconstruction attack for privacy preservation~\cite{wu2021fedcg}. Recently, GAN has been used to establish secure FL by using GAN parameters instead of sharing classifier parameters to avoid privacy leakage since GAN holds encrypted domain distribution~\cite{ma2024ppidsg}. To ensure security against attacks, researchers commonly employ DP-based methods~\cite{abadi2016deep,liao2023ppgencdr,wei2020federated} or follow gradient pruning, gradient perturbation-based methods~\cite{zhu2019deep,sun2021soteria}. These methods sacrifice performance to resist privacy leakage due to non-systematic gradient alteration in the form of a defense mechanism, which also affects the overall learning of the FL setup.

Considering the aforementioned limitations, our approach improves the utility-security trade-off in FL. To achieve this, we propose a stochastic bidirectional learning approach that allows generalized learning in local clients through diverse updates/ models from the global server such that these diverse solution updates are in close neighborhoods. Using diverse but close neighborhood updates, the clients follow generalized solutions and hence improve the classification accuracy to improve the utility of FL. To ensure security without hurting utility, our approach follows systematic updates in gradients of the diverse solutions sent from the global server to different clients so that updates are optimally closer.

\section{Preliminaries}
\subsection{Overview of Federated Learning} FL follows a cloud-server architecture, which consists of a global server and multiple local clients. In this paper, we consider the FL system to consist of homogenous local client models, i.e., using similar data distribution and having a model structure as of the global model. Assume our FL aims to map input space $\mathcal{X}$ to output space $\mathcal{Y}$ using global model, $\mathcal{S}$ and $\mathcal{N}$ local models represented as \{$c_1$, $c_2$, \dots, $c_N$\}. We denote the local dataset for each client $i$ as $\mathcal{D}_i$ = \{$x_{i,1}$, $x_{i,2}$, \dots, $x_{i,n_i}$\} such that $x_{i,j}$ = $(\mathcal{X}, \mathcal{Y}) \in \mathcal{X} \times \mathcal{Y}$. In FL, global model weights $w_{glb}$ are trained collaboratively by local clients by sharing their learning (local models) with the global server. The conventional method to generate a global model by aggregating the local models is FedAvg~\cite{bonawitz2017practical}. We can formulate the primary objective of FL using Eq.~\ref{eq:FL_Objective}. 
\begin{equation}\label{eq:FL_Objective}
\begin{split}
     \underset{w}{\text{min}}~~ \mathcal{L} (w) &= \frac{1}{N} \sum_{i=1}^N f_i(w), \\
     \text{where}, \quad f_i(w) &= \frac{1}{n_i} \sum_{j=1}^{n_i} l(w;x_{i,j})
\end{split}
\end{equation}
where, $\mathcal{L}$ denotes the loss terms for the global model, $l$ denotes loss for individual samples, $f_i$ comprises the loss of all samples for local client $i$.

\subsection{Privacy Leakage Using Attacks in FL} In our work, we hold the assumption that the adversary does not corrupt the training. The client models share their parameters (weights or gradients) with the global model. When all local clients train their models for just one local epoch between two global aggregation operations, using their complete training datasets, we assume these parameters are equivalent. This scenario results in a white-box attack where model parameters and structure are accessible to the adversary.\\
\noindent
\textbf{Label Inference Attack (LIA):} Assume each client $C_k$ with local dataset $\mathcal{D}_k = \{ (x_i, y_i) \}_{i=1}^{n_k}$ where $x_i$, $y_i$ denotes the $i$-th sample and its ground truth label respectively. Consider local training with batch size, $bs$ with cross-entropy loss, $\mathcal{L}$ for the classification task; we can define the gradient of $\mathcal{L}$ with respect to (wrt) $w$ using Eq.~\ref{eq:LIA-eq1}
\begin{equation}\label{eq:LIA-eq1}
    \nabla_{w} \mathcal{L}(\mathbf{x}, \mathbf{y})=-\frac{1}{b s} \sum_{i=1}^{bs} \sum_{j=1}^{n_c}\nabla_{w}[y_i(j)\log y_i'(j)]
\end{equation}
where, $n_c$ denotes the number of class labels, $y_i'$ denotes the predicted logit. $y_i(j)$=1 if output index $j$ matches ground truth else $y_i(j)$=0. In LIA, our aim is to count images present in a batch ($\sum_{i=1}^{bs} y_i(j)$) for each class type $j$. For each input $x_i$, we can compute gradient wrt network output $z_i$ at index $j$ as suggested in (Yin et al. 2021) as shown in Eq.~\ref{eq:LIA-eq2}: 
\begin{equation}\label{eq:LIA-eq2}
    \nabla_{z_i(j)} \mathcal{L}(\mathbf{x_i}, \mathbf{y_i})=y_i'(j)-y_i(j)
\end{equation}
Further, we can use the uploaded gradient from the classifier model to perform LIA by multiple passes of random samples to the classifier to compute each category count $j$, which leads to privacy leakage~\cite{ma2024ppidsg}. We do not share the classifier parameters, so our method is resistant to LIA. 
\noindent
\textbf{Membership Inference Attack (MIA):}
In our work, we consider the enhanced MIA~\cite{ma2024ppidsg}, defined as follows: Suppose attacker $C_{adv}$ has shadow dataset $\mathcal{D^*}$ which contains some images with target distribution but not in the target dataset. To improve attacks, shadow models are rebuilt for victims and other users. Using model parameters from clients, $C_{adv}$ generates a copy of victim model ${M^{victim}}$ and other models ${M^{others}}$ (aggregate if other users $>$ 2). The adversary $C_{adv}$ produces non-overlapping datasets randomly using $\mathcal{D^*}$ as $\mathcal{D}^*_{\text{victim}}$ and $\mathcal{D}^*_{\text{others}}$, which are fed as input to${M^{victim}}$, ${M^{others}}$ respectively. The obtained predictions ${P^*_{victim}}$, ${P^*_{others}}$ are manually considered with labels as $in$ (member) and $out$ (non-member) respectively. Using obtained predictions and their labels, an inference model ${M_{attack}}$ is trained by the adversary. Since the adversary has a skeptical dataset, it becomes difficult to determine whether the data comes from the victim or other users. $C_{adv}$ fed this dataset to ${M_{attack}}$ to perform MIA to infer data samples, and its success is determined based on correct inferences.\\
\noindent
\textbf{Image Reconstruction Attack (IR):}
During IR, the adversary attempts to recover the original image through the encrypted image to perform privacy leakage. To achieve this, optimization aimed to minimize the gradient difference obtained through the dummy images $x_{dummy}$ having labels $y_{dummy}$ and gradients shared by victim $\nabla w$. We can formally denote IR using Eq.~\ref{eq:IR_Attack}.
\begin{equation}\label{eq:IR_Attack}
    x_{dummy}=\arg  \underset{x} {min} \left\|\frac{\partial \mathcal{L}\left(x, y_{dummy} ; w\right)}{\partial w}-\nabla w\right\|^2
\end{equation}

%%%%%%%%%%%%%%%%%%%%%%%%%%%%%%%%%%%%%%%%%%%%%5

\section{Methodology} 
We propose a stochastic bidirectional parameter update mechanism to improve the utility of clients as well as improve their defense against different attacks. To achieve this, our approach generates diverse and generalizable global models through systematic perturbations using a dual gradient mechanism such that the diverse global models for different clients are in close neighborhoods. Our approach improves the robustness of clients against different attacks without sacrificing model utility.

To validate the effectiveness of our method, we followed the setup similar to PPIDSG~\cite{ma2024ppidsg} and did not share the classifier ($C$) parameters and used GAN to share the parameters from generator ($G$) parameters in FL, where $G$ learns the image distribution in the encrypted domain. We augment training data and encrypt it by utilizing the image distribution scheme proposed by PPIDSG, where we encrypt training images through several transformations like Rotation and augmentation using a pseudo-random bit, image block flipping, and pixel value exchange across channels. We learn target distribution in $G$ by adversarial training with Discriminator, $D$. We used auto-encoder~\cite{sellami2022deep} to build a feature extractor $F$ and used a separate classifier network $C$ to train it. The obtained classifier loss is fed to $G$ to help in learning class specific distributions to improve classification.

Our FL setup consists of $K$ clients ($C_k$) having their local dataset $\mathcal{D}_\text{k}$ with $N$ samples in total. During $FL$, clients update their model locally and share the generator parameters $w$ with the central/global server. The central server takes updates sent from the local clients and aggregate model parameters as shown in Eq.~\ref{eq:aggregate}, where $\alpha$ and $\mathcal{L}$ denote the learning rate and loss function, respectively.

\begin{equation}\label{eq:aggregate}
w^{t+1} \leftarrow \sum_{k=1}^{K} \frac{n_k}{N}\left( w^t - \alpha \nabla L(w^t; D_k) \right)
\end{equation}

To generate a diverse global model for each client, we proposed a Stochastic Bidirectional Parameter Updates ($SBPU$) strategy, which utilizes the global models from previous FL rounds. We elaborate on our proposed approach as follows:

\begin{figure}
    \centering
    \includegraphics[width=8cm, height=4.2cm]{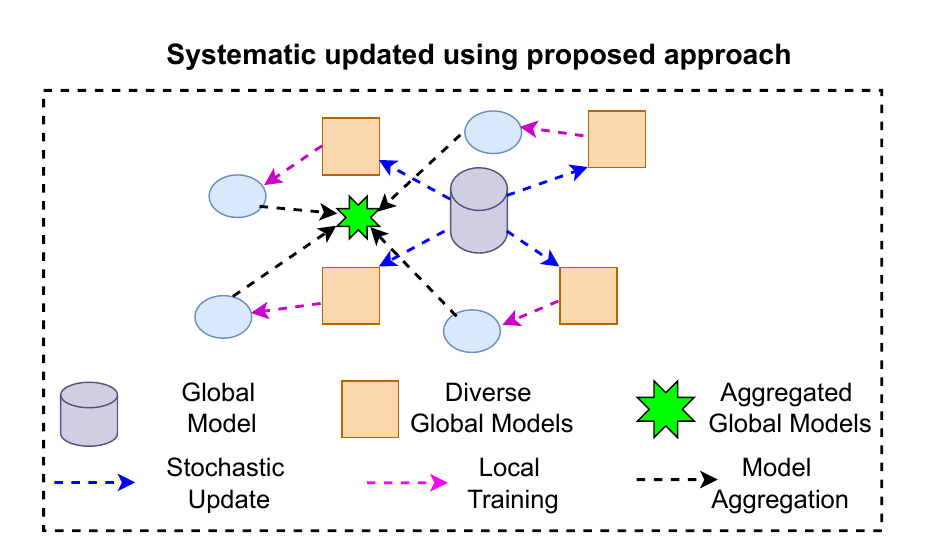}
    \caption{Visualization of diverse global models obtained using proposed approach ($SBPU$). We generate diverse models in close neighborhoods through systematic updates. It is important to note that we do not make perturbations after generating the diverse models, which helps to retain utility and generalization. }
    \label{fig:generalized_soution}
\end{figure}

\subsection{Stochastic Bidirectional Parameter Updates using Dual-Gradient Mechanism}\label{SBPU_Detailed}
Our novelty lies in proposing generalized updates for local clients to improve the utility and resist privacy leakage in FL. We validated the superiority of our method over the recent state-of-the-art method~\cite{ma2024ppidsg} (PPIDSG) by following a similar FL set-up (except using the proposed approach to generate diverse global solutions) and a further improvement in model utility and defense across the attacks. To achieve this, we propose a stochastic bidirectional learning mechanism that helps to generate diverse solutions at the global server to update local clients. The generated models at the server are in the neighborhood to provide generalized solutions for FL clients (refer to Fig.~\ref{fig:generalized_soution}). 

\begin{figure*}
    \centering
    \includegraphics[trim=1.2cm 0.5cm 0.5cm 0.3cm, clip, width=16.27cm, height=4.65cm]{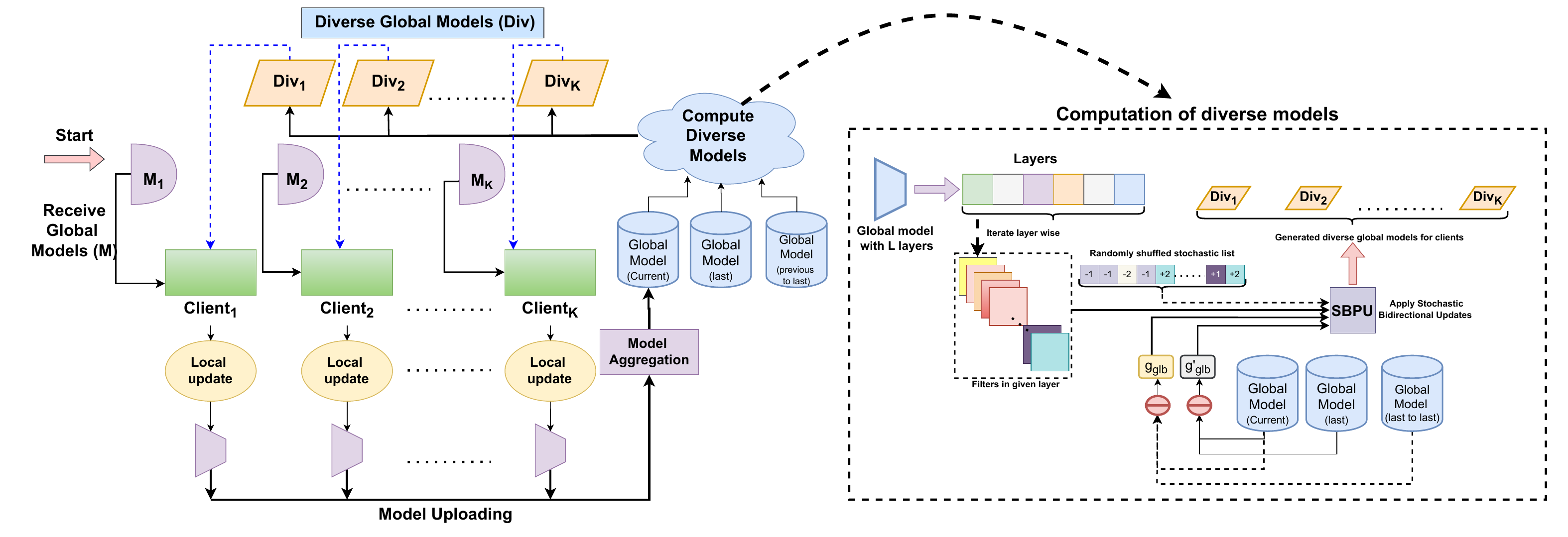}
    \caption{The schematic architecture diagram of the proposed approach.} % Our Proposed Architecture
    \label{fig:proposed_architecture}
\end{figure*}

Our approach makes bidirectional systematic alterations in the gradients by modification in model parameters at a fine-grained level (i.e., altering each convolutional filter across the layers of the model), which improves the defense against attacks. These systematic updates provide a diverse model for each client and help them with generalized learning to improve the utility of the model (refer Fig.~\ref{fig:generalized_soution}). The overview of the proposed method is provided in Fig.~\ref{fig:proposed_architecture}.

The overall FL round mainly consists of four steps: 1) The global model sends a diverse model to each client, as shown in Algo.~\ref{algo:dual_grad_compute}. 2) With the received diverse model, each client undergoes local training and then uploads the model parameters (in our case, $G$) to the global model. 3) The global model aggregates the received models from the clients as shown in Eq.~\ref{eq:aggregate} to obtain the updated global model. 4) Using the current global model and previous global models, we perform Stochastic Bidirectional Parameter Updates ($SBPU$) to generate diverse models as shown in Algo.~\ref{algo:SBPU}. For each client, the global model sends one diverse model to assist in its learning in the next round. Finally, we update the global models ($w_{glb}$, $w'_{glb}$, $w''_{glb}$) for the next stochastic update.

Please note that, for the initial two FL rounds, we initialize $w'_{glb}$ and $w''_{glb}$ with $w_{glb}$.To perform $SBPU$, we use $StochasticList$ as [-1, $\dots$, -1, 1, $\dots$, 1, -2, $\dots$, -2, 2, \dots, 2], where the frequency of each stochastic term, i.e., \{-1, 1, -2, 2\}, equals to $\left\lfloor \frac{f}{4} \right\rfloor$ where $f$ denotes the number of filters present in layer $i$ of global model $w_{glb}$. The $StochasticList$ is randomly shuffled to create diverse models by performing bidirectional parameter updates in filter $j$ from layer $i$. If $StochasticList[j] == \pm1$, we perform the update under diversity rate $\beta_1$ as shown in Line 9 of Algo.~\ref{algo:SBPU}, else we update under diversity rate $\beta_2$ as shown in Line 11 of Algo.~\ref{algo:SBPU}, where $g_{glb}.i.j$, $g'_{glb}.i.j$ denotes the gradient computed for $j$-th filter of $i$-th layer using global models from previous and previous to previous FL rounds, respectively; and $w_{loc}.i.j$ denotes the diverse model obtained after $SBPU$ for $j$-th filter of $i$-th layer.

\begin{algorithm}
\caption{Diverse Models Generation}\label{algo:dual_grad_compute}
\textbf{Input:} 
i) $w_{glb}$, the global model, ii) $w'_{glb}$, the previous global model, iii) $w''_{glb}$, the previous to previous global model, \\
iv) $K$, \# clients for FL. \\
\textbf{Output:} 
i) $w_{diverse}$, the list of diverse models \\
\textbf{Intermediate:} ($w_{glb}, w'_{glb}, w''_{glb}$)
\begin{algorithmic}[1]
\State $g_{glb} \leftarrow w_{glb} - w'_{glb}$ \Comment{gradients using previous model}
\State $g'_{glb} \leftarrow w_{glb} - w''_{glb}$ \Comment{gradients using previous to previous model}
\For{$i \leftarrow 1, \ldots, K$}
    \State $w_{diverse}.\text{append}(SBPU(w_{glb}, g_{glb}, g'_{glb}))$
\EndFor
\State \Return $w_{diverse}$
\end{algorithmic}
\end{algorithm}

\begin{algorithm}
\caption{Stochastic Bidirectional Parameter Updates ($SBPU$)}\label{algo:SBPU}
\textbf{Input:} 
i) $w_{glb}$: global model, ii) $g_{glb}$: gradients using previous run, iii) $g'_{glb}$: gradients using previous to previous run, iv) $\beta_1$, $\beta_2$: diversity rates \\
\textbf{Output:} 
i) $w_{loc}$, the mutated weights \\
\textbf{$SBPU$}($w_{glb}, g_{glb}, g'_{glb}$)
\begin{algorithmic}[1]
\State $w_{loc} \leftarrow$ $copy(w_{glb})$
\State $L \leftarrow$ the number of layers of $w_{glb}$
\For{$i \leftarrow 1, \ldots, L$}
    \State $f \leftarrow w_{glb}[i].\text{size}$ \Comment{number of filters}
    \State ${StochasticList} \leftarrow \left[ -1, -1, \ldots, -1, 1, 1, \ldots, 1 , -2, -2, \ldots , -2, 2, 2, \ldots, 2 \right]$ \Comment{each type of stochastic element counts to $\left\lfloor \frac{f}{4} \right\rfloor$}
    \State $StochasticList \leftarrow \text{Shuffle}(StochasticList)$ %\Comment{shuffle the stochastic list}
    \For{$j \leftarrow 1, \ldots, f$} 
        \If{$StochasticList[j] == \pm1$} 
            \State $w_{loc}[i][j] \leftarrow w_{glb}[i][j] + \beta_1 \times {StochasticList}[j] \times g_{glb}[i][j]$  
        \Else
            \State $w_{loc}[i][j] \leftarrow w_{glb}[i][j] + \beta_2 \times {StochasticList}[j] \times g'_{glb}[i][j]$ 
        \EndIf
    \EndFor
\EndFor
\State \Return $w_{loc}$
\end{algorithmic}
\end{algorithm}

\subsection{Overall Architecture}\label{sec:PPIDSG_Detailed}
In the GAN setup, $G$ captures the distribution in encrypted space and shares its parameters to perform FL. The original image is fed to $G$ instead of noise as it improves privacy. The generator network, $G$, consists of the encoder, ResNet blocks, and a decoder. To extract features from the original image, the encoder is used, which is fed into the ResNet block to maintain and align the image features into the target domain. Finally, the decoder helps to restore the features of the image. 
The discriminator network $D$ utilizes the adversarial loss (without conditional labels), $\mathcal{L}_{\text{adv}}$ for effective conversion into the encrypted domain. Consider original image domain ($X$) with distribution $x_i \sim p_{\text{data}}(x)$ and target domain $\hat{X}$ with distribution $\hat{x_i} \sim p_{\text{data}}(\hat{x})$. For training images with batch size $bs$, we can express the objective using Eq.~\ref{eq:adver_loss} where $G$ and $D$ tires to maximize and minimize the objective, respectively. 
\begin{align}\label{eq:adver_loss}
    \mathcal{L}_{\text{adv}} = &\ \mathbb{E}_{\hat{x}_i \sim p_{\text{data}} (\hat{x})} \left[ \log D\left( \{\hat{x}_i\}_{i=1}^{b_s} \right) \right] \notag \\
    &+ \mathbb{E}_{x_i \sim p_{\text{data}} (x)} \left[ \log \left( 1 - D \left( G \left( \{x_i\}_{i=1}^{b_s} \right) \right) \right) \right]
\end{align}
To retain semantic information, we use semantic loss using $l_1$ norm as shown in Eq.~\ref{eq:l1_norm}, where $\theta_G$ denotes the generator parameters.
\begin{equation}\label{eq:l1_norm}
    \mathcal{L}_{\text{sem}} = \sum_{i=1}^{bs} \left\| G_{X \rightarrow \hat{X}} (x_i; \theta_G) - \hat{x}_i \right\|
\end{equation}
\begin{figure}
    \centering
    \includegraphics[width=8.8cm, height=3.64cm]{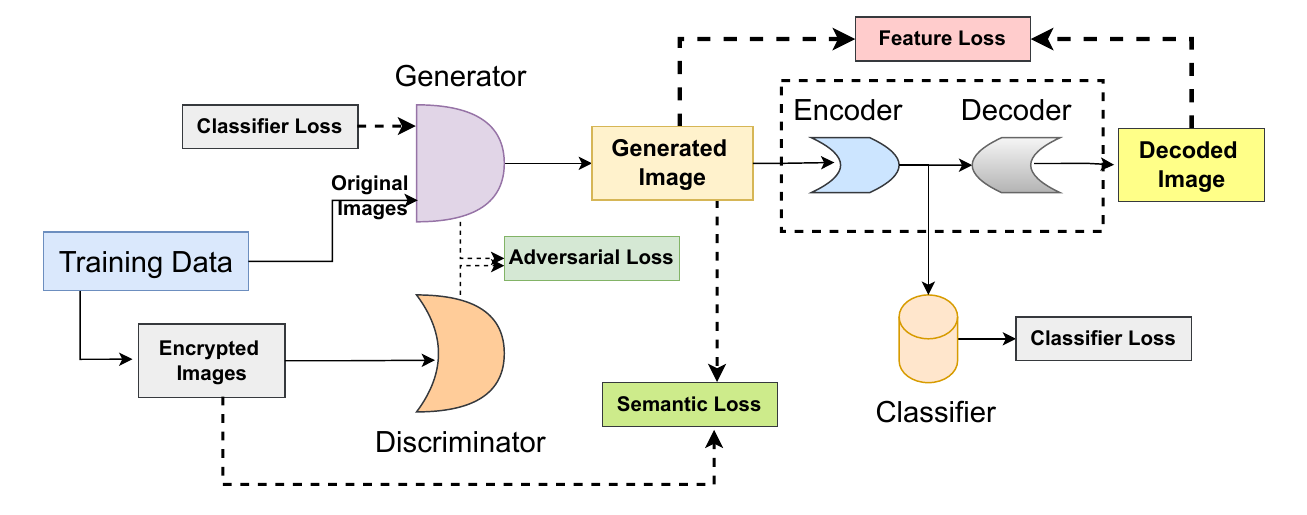}
    \caption{High-level representation of different components in the overall architecture.}
    \label{fig:enter-label}
\end{figure}
To improve distribution learning in $G$ towards the class-specific features, we used classification loss $\mathcal{L_\text{cls}}$ into $G$. With training epochs, $G$ learns to align generated images into the encrypted domain, and then generator parameters can be shared to the global server by different clients to facilitate FL. To improve the feature learning, we use feature extractor $F$, which consists of Encoder ($Enc$) and decoder $Dec$ networks. The features extracted by $Enc$ are converted into images using $Dec$. To extract efficient features, $F$ tries to minimize the feature distance between the image generated by $G$ and $Dec$ as shown in Eq.~\ref{eq:classifer} where $\tilde{x_i}$ denotes the generated image. The classifier $C$ consists of a simple convolutional network and takes features from $F$ to minimize the classification error ($\mathcal{L_\text{cls}}$) for $n_c$ classes as described in Eq.~\ref{eq:clf_loss}. We compute total loss $\mathcal{L}_{\text{total}}$ as shown in Eq.~\ref{eq:overall_loss} where $\lambda_{sem}$ and $\lambda_{cls}$ are hyperparameters to control the influence of $\mathcal{L}_{\text{sem}}$ and $\mathcal{L}_{\text{cls}}$ respectively.
\begin{equation}\label{eq:classifer}
    \mathcal{L}_{\text{fea}} = \sum_{i=1}^{bs} \left\| \text{Dec}(\text{Enc}(\tilde{x}_i)) - \tilde{x}_i \right\|^2
\end{equation}
\begin{equation}\label{eq:clf_loss}
    \mathcal{L}_{\text{cls}} = - \sum_{i=1}^{bs} \sum_{j=1}^{n_c} y_i(j) \log y'_i(j)
\end{equation}
\begin{equation}\label{eq:overall_loss}
    \mathcal{L}_{\text{total}} = \mathcal{L}_{\text{adv}} + \lambda_{sem} ~ \mathcal{L}_{\text{sem}} + \lambda_{cls} ~ \mathcal{L}_{\text{cls}}
\end{equation}

% ***********************************************************************************
\subsection{Convergence Analysis}
Our global model is aggregated from all the trained local models similar to FedAvg. Let $t$ denote the $t^{th}$ SGD iteration on the local client, and each local client undergoes $E$ SGD training iterations, $w_{glb}$ denotes the aggregated model. For $i = 1, 2, \dots, K$, our method satisfies the following property:
\begin{equation}
\alpha^2 \|{w}_{{glb}}^n - {w}_{{glb}}^{n-1}\|^2 \le  \|{w}_{{i,n}}^{loc} - {w}_{{glb}}^{n}\|^2 \le 4\alpha^2 \|{w}_{{glb}}^{n} - {w}_{{glb}}^{n-1}\|^2
\end{equation}
where $\mathbf{w}_{i,n}^{{loc}}$ denotes the $i^{th}$ mutated weights in the $n^{th}$ round. Inspired by~\cite{li2019convergence}, the following assumptions on the loss functions of local clients (i.e., $f_1, f_2, \dots, f_K$) can be considered.\\
\textbf{Assumption 1:} For $i \in \{1, 2, \dots, K\}$, $f_i$ is $L$-smooth, where $f_i(v) \leq f_i(w) + (v - w)^T \nabla f_i(w) + \frac{L}{2} \|v - w\|_2^2$.\\
\textbf{Assumption 2:} For $i \in \{1, 2, \dots, K\}$, $f_i$ is $\mu$-strongly convex, where $f_i(v) \geq f_i(w) + (v - w)^T \nabla f_i(w) + \frac{\mu}{2} \|v - w\|_2^2$.\\
\textbf{Assumption 3:} The variance of stochastic gradients is bounded by $\sigma_i^2$, i.e., $\mathbb{E}\|\nabla f_i(w; \xi) - \nabla f_i(w)\|^2 \leq \sigma_i^2$, where $\xi$ is a data batch of the $i^{th}$ client in the $t^{th}$ FL round.\\
\textbf{Assumption 4:} The expected squared norm of stochastic gradients is bounded by $G^2$, i.e., $\mathbb{E}\|\nabla f_i(w; \xi)\|^2 \leq G^2$.\\
Based on these assumptions, our convergence can be obtained as:\\
\textbf{Theorem 1. (Convergence of SBPU)} Let Assumption 1-4 hold. If there are $n$ FL rounds during the FL training process. Let $T = n \times E$ denotes total number of SGD iterations and $\eta_t = \frac{2}{\mu (t + \gamma)}$ is learning rate. Let $\kappa = \frac{L}{\mu}$, $\gamma = \max(8\kappa, E)$. We have
\begin{equation}
\mathbb{E}[f({w}_T) - f^*] \leq \frac{4 \kappa}{\gamma + T} (\frac{B}{2\mu} + L \|{w}_1 - {w}^*\|^2)
\end{equation}
where $B = \frac{1}{K^2} \sum_{i=1}^K \sigma_i^2 + \frac{32 \alpha^2}{1-4\alpha^2}(E - 1)^2 G^2$. Theorem 1 computes loss for SBPU between $f^*$ (optimal weight) and $f({w}_T)$ in the $T^{th}$ interaction and indicates a convergence rate similar to FedAvg (detailed in~\cite{li2019convergence}). We have provided full proof of convergence analysis of SBPU in the supplementary material.

% Theorem 1 shows loss difference in $T^{th}$ interactions, i.e., $f({w}_T)$, and the optimal loss, i.e., $f^*$. This result shows SBPU holds a convergence rate similar to that of FedAvg, detailed in~\cite{li2019convergence}. Full proof related to the convergence analysis of SBPU is provided in the supplementary material.

\section{Experimental Setup and Results Obtained}\label{sec:experimental_results}
\subsection{Dataset Used and Implementation Setup}
We implemented the proposed approach using the PyTorch programming framework using NVIDIA RTX A5000 GPU with 24GB GPU memory. We evaluated our model on four datasets MNIST~\cite{deng2012mnist}, FMNIST~\cite{xiao2017fashion}, CIFAR10~\cite{krizhevsky2009learning}, and SVHN~\cite{yuval2011reading} as per official train-test split. To perform different attacks, we randomly select any one client as a victim.  

\subsection{Implementation Details}
For a fair comparison with recent benchmark PPIDSG~\cite{ma2024ppidsg}, we followed a similar setting, i.e., homogeneous distribution across clients in the FL system~\cite{mcmahan2017communication} having 10 clients with equal training data access. We followed a batch size of 64 for GAN. We used block sizes ($B_x$ and $B_y$) for image encryption as 4. The generator and discriminator network have been trained using Adam optimizer with a learning rate (lr) of 0.0002. For the feature extractor ($F$) and classifier ($C$), we used an SGD optimizer with an lr of 0.01 and weight decay of 0.001. We keep the initial lr constant for the first 20 global iterations and then follow linear decrement until it converges to 0. In Eq.~\ref{eq:overall_loss}, we used $\lambda_{sem}$=1 and $\lambda_{cls}$=2 and train model for 100 rounds. We used $\beta$ = 0.025, 0.25, 0.15, and 1.1 in our learning rule for MNIST, FMIST, CIFAR10, and SVHN datasets, respectively. We defined $\beta_1$ and $\beta_2$ as $\beta$ and $\beta^2$, respectively. For further details, please refer to the supplementary material.

\subsection{Defense Baselines}
To evaluate the robustness of our model against different attacks, we compared it against several defense mechanisms, i.e., 1) ATS~\cite{gao2021privacy} (to find optimal image transformation through automatic transformation search) 2) EtC~\cite{chuman2018encryption} (encrypt image using block-based image transformation) 3) DP~\cite{wei2020federated} (use clipped gradient with gaussian noise during model training) 4) GC~\cite{zhu2019deep} follows gradient pruning to avoid privacy leakage 5) FedCG~\cite{wu2021fedcg} use conditional GANs for privacy preservation in FL 6) PPIDSG~\cite{ma2024ppidsg} (share GAN parameters during FL rather than classifier to ensure privacy preservation). For differential privacy (DP), we kept the privacy budget as $\epsilon/T$ for $T$ global training epochs with clipping hyperparameter $C$ and denoted as DP $<$$\epsilon,C$$>$.

% \subsection{Evaluating Utility and Robustness Against Attacks:}
\subsection{Utility and Robustness Against Attacks}
To validate the utility of the model, we select a random user to evaluate classification performance since our approach doesn't have a global classification model. We provide the highest classification accuracy obtained by our model and compare it with different techniques in Table~\ref{tab:classification_accuracy}. Our method surpassed the state-of-the-art (SOTA) methods and obtained the highest classification accuracy. For CIFAR10, we observed significant improvement in classification accuracy. It is important to note that ATS and EtC defense policies utilize ResNet-18 as a classifier, which mainly determines the model utility and affects the classification accuracy. Our model uses a simple classifier network and is able to improve model utility mainly due to diverse and generalized updates to clients. To ensure effectiveness against attacks, our approach does not share the classifier parameter. Since we do not share the classifier model, our model becomes robust against Label Inference Attack (LIA). We have shown the robustness of other defense methods against LIA using different activation functions and architecture in the supplementary material. We considered enhanced Membership Inference Attack (MIA) as proposed in~\cite{ma2024ppidsg} for evaluation. We compared our defense accuracy against MIA with other defense mechanisms in Table~\ref{tab:mia_cifar_svhn}. 

\begin{figure*}
    \centering
    \includegraphics[width=17.2 cm, height=5.2cm]{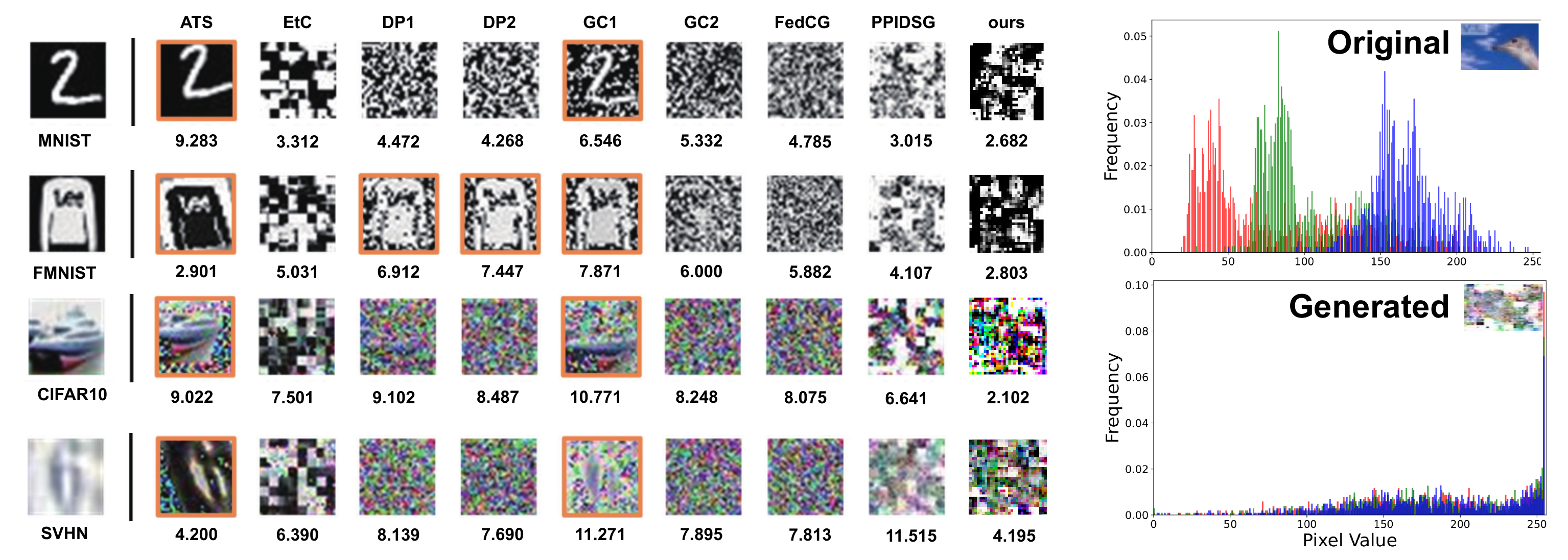}
    \caption{Comparative analysis under reconstruction attack. Here, a small PSNR value denotes privacy preservation, i.e., the robustness of the model against IR attack. Through histogram plots, we can see that the generated image for a given CIFAR10 sample is encrypted and does not reveal visual information (either visually or through pixel distribution), which affirms the effectiveness of our method.}\label{fig:RA_attack}
\end{figure*}

Due to page limit constraints, we provided additional results, such as model utility and its robustness against MIA under $Part$ and $All$ settings for the remaining datasets in the supplementary material.  For Reconstruction Attack (RA), we do not share the classifier model parameters with the global model, so the adversary fails to perform RA on our approach and attempts to reconstruct the image using generator parameters. In Fig.~\ref{fig:RA_attack}, we provide the quantitative and qualitative comparison of different mechanisms with the proposed approach against RA.

\begin{figure}
    \centering
    \includegraphics[width= 7.6cm, height = 3cm]{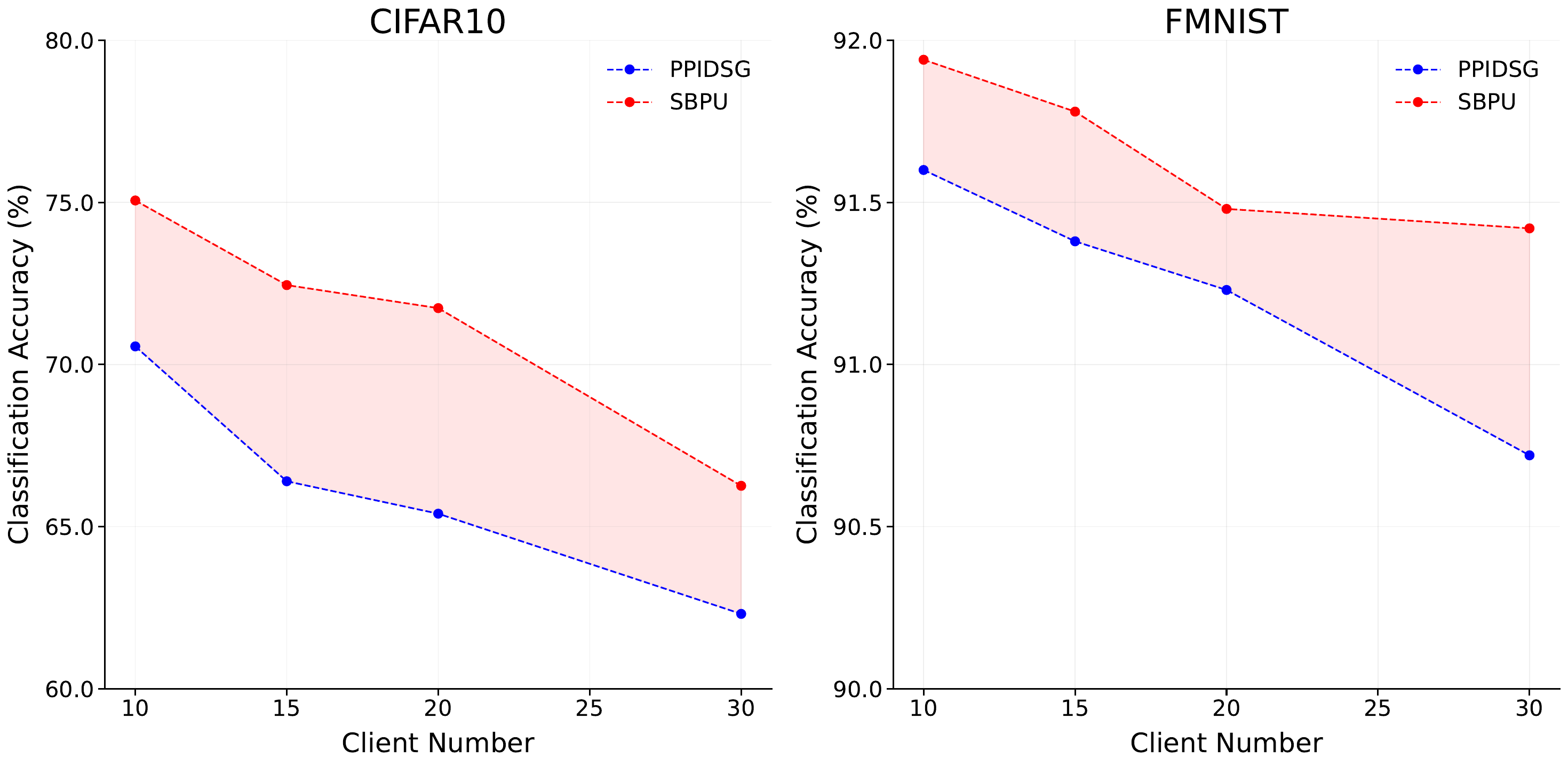}
    \caption{Comparison with PPIDSG using CIFAR10 and FMNIST.}
    \label{fig:client_compare_CIFAR-10}
\end{figure}

\begin{table}
\centering
\fontsize{9}{10}\selectfont
\begin{tabular}{ccc}
\toprule
\textbf{Method} & \textbf{CIFAR10 $\uparrow$} & \textbf{SVHN $\uparrow$} \\ \midrule
ATS    & 59.67 & 85.22 \\ 
EtC    & 53.34 & 78.7  \\ 
DP1    & 49.29 & 82.7  \\ 
DP2    & 44.43 & 80.28 \\ 
GC1    & 54.07 & 84.36 \\ 
GC2    & 50.91 & 79.96 \\ 
FedCG  & 53.2  & 79.71 \\ 
PPIDSG & 70.56 & 91.53 \\ \midrule
\textit{\textbf{Ours}}   & \textit{\textbf{75.06}} & \textit{\textbf{92.36}} \\ \bottomrule
\end{tabular}
\caption{Comparison of classification accuracy (\%) obtained for different policies on various datasets. Here, DP1: DP$<$5,10$>$, DP2: DP$<$20,5$>$, GC1: GC (10\%), GC2: GC (40\%).} \label{tab:classification_accuracy}
\end{table}

\begin{table}
\centering
\fontsize{9}{10}\selectfont
\begin{tabular}{ccccc}
\toprule
\multirow{2}{*}{\textbf{Method}} & \multicolumn{2}{c}{\textbf{CIFAR10}} & \multicolumn{2}{c}{\textbf{SVHN}} \\ \cmidrule{2-5}
 & \textbf{Part $\downarrow$} & \textbf{All $\downarrow$} & \textbf{Part $\downarrow$} & \textbf{All $\downarrow$} \\ \midrule
ATS    & 84.3  & 73.51 & 54.78 & 52.41 \\ 
EtC    & 55.84 & 46.83 & 50.04 & 63.89 \\ 
DP1    & 69.6  & 68.74 & 56.37 & 63.55 \\ 
DP2    & 70.28 & 65.95 & 59.37 & 55.76 \\ 
DP3    & 87.17 & 75.47 & 62.21 & 55.19 \\ 
DP4    & 73.71 & 56.44 & 59.31 & 57.53 \\ 
DP5    & 74.29 & 66.64 & 58.66 & 58.66 \\ 
GC1    & 75.56 & 72.43 & 60.74 & 58.31 \\ 
GC2    & 67.41 & 61.18 & 56.37 & 56.01 \\ 
GC3    & 83.71 & 84.16 & 58.13 & 54.24 \\ 
FedCG  & 49.84 & 70.03 & 51.18 & 53.72 \\ 
PPIDSG & 54.39 & 52.54 & 52.35 & 47.21 \\ \midrule
\textbf{\textit{Ours}}   & \textbf{\textit{40.31}} & \textbf{\textit{36.11}} & \textbf{\textit{49.44}} & \textbf{\textit{39.52}} \\ \bottomrule
\end{tabular}
\caption{The MIA accuracy (\%) obtained for different defense mechanisms on CIFAR10 and SVHN dataset under $Part$ and $All$ settings. Here, lower accuracy shows the robustness of the model in resisting the privacy leakage against the MIA. Here, DP1, DP2, DP3, DP4, DP5, GC1, GC2, and GC3 denote DP$<$5,10$>$, DP$<$10,10$>$, DP$<$20,10$>$, DP$<$20,5$>$, DP$<$20,20$>$, GC(10\%), GC(20\%), and GC(40\%).}\label{tab:mia_cifar_svhn}
\end{table}

\setlength{\tabcolsep}{3pt}
\begin{table}
\centering
\fontsize{9}{10}\selectfont
\begin{tabular}{cccccc}
\toprule
\textbf{DatatSet} & \textbf{$\beta$} & \textbf{SingleGrad}  & \textbf{DualGrad} & \textbf{TripleGrad} \\
\midrule
\multirow{4}{*}{CIFAR10} & 0.15 & 0.17(0.62) & 0.17(0.54) & 0.19(0.59) \\
 & 0.25 & 0.15(0.66) & 0.15(0.56) & 0.20(0.60) \\
 & 0.5 & 0.16(0.71) & 0.16(0.59) & 0.10(0.67) \\
 & 1.5 & 0.16(0.67) & 0.15(0.67) & 0.24(0.78) \\
 \bottomrule
\end{tabular}
\caption{Ablation study to decide stochastic bidirectional update mechanism under different global models consideration. The values present outside and inside the brackets denote the F1-score obtained by member and non-member, respectively.}
\label{tab:ablation_beta}.
\end{table}

% fmnist 0.00027896904945373533 0.0002593759536743164
% mnist  0.0002581131458282471 0.0002643908739089966
% svhn 0.00026313682128552214 0.0002562086422456578
% cifar 0.0002648900477667857 0.0002544744749631923

% \setlength{\tabcolsep}{3pt}
\begin{table}[h!]
\centering
% \begin{adjustbox}{width=0.47\textwidth}
\fontsize{9}{10}\selectfont
\begin{tabular}{ccccc}
% ***
\toprule
\multirow{2}{*}{\textbf{Dataset}} & \multicolumn{2}{c}{\textbf{Train (seconds)}} & \multicolumn{2}{c}{\textbf{Test ($\times 10^{-4}$ seconds)}} \\ \cmidrule{2-5}
 & \textbf{PPIDSG} & \textbf{Proposed} & \textbf{PPIDSG} & \textbf{Proposed} \\ \midrule
% ****
% \hline
% \textbf{Datasets} & \textbf{PPIDSG (train)} & \textbf{Proposed (train)} & \textbf{PPIDSG (test)} & \textbf{Proposed (test)} \\ \hline
MNIST   & 111.76 & 112.90 & $2.59$ & $2.62$ \\ 
CIFAR10 & 123.52 & 125.26 & $2.58$ & $2.65$ \\ 
SVHN    & 186.73 & 187.99 & $2.56$ & $2.64$\\ 
FMNIST  & 111.30 & 111.79 & $2.54$ & $2.60$ \\ \bottomrule
\end{tabular}
% \end{adjustbox}
\caption{Comparison of training time (per epoch) and testing time (per image) for PPIDSG and the proposed method (SBPU) across different datasets.}
\label{tab:training_times}
\end{table}

% \subsection{Ablation Study} 
% To decide the proposed stochastic bidirectional parameter update strategy, we performed ablation to analyze the effect of varying the number of global models (one, two, three, i.e., SingleGrad, DualGrad, and TripleGrad) to consider from previous FL rounds. SingleGrad, DualGrad, and TripleGrad consider stochastic terms for $StochasticList$ as \{$\pm$ 1\} with diversity rate: $\beta$; \{$\pm$ 1, $\pm$ 2\} with diversity rate: $\beta$, $\beta^2$; and \{$\pm$ 1, $\pm$ 2, $\pm$ 3\} with diversity rate: $\beta$, $\beta^2, \beta^3$, respectively. 
% We provided these ablations in Table~\ref{tab:ablation_beta}.
% We performed ablation to analyze the effect of i) block size for image encryption, ii) the number of clients, and iii) the Effect of varying diversity rates, iv) compared the defense accuracy of
% our approach with SOTA settings: \textit{original image} and \textit{no update}; and provided results in the supplementary material.

\subsection{Ablation Study} 
To decide the proposed stochastic bidirectional parameter update strategy, we performed ablation to analyze the effect of varying the number of global models (one, two, three, i.e., SingleGrad, DualGrad, and TripleGrad) to consider from previous FL rounds. SingleGrad, DualGrad, and TripleGrad consider stochastic terms for $StochasticList$ as \{$\pm$ 1\} with diversity rate: $\beta$; \{$\pm$ 1, $\pm$ 2\} with diversity rate: $\beta$, $\beta^2$; and \{$\pm$ 1, $\pm$ 2, $\pm$ 3\} with diversity rate: $\beta$, $\beta^2, \beta^3$, respectively. We provided these ablations in Table~\ref{tab:ablation_beta}.
We performed ablation to analyze the effect of i) block size for image encryption, ii) the number of clients, and iii) the Effect of varying diversity rates, iv) compared the defense accuracy of our approach with SOTA settings: \textit{original image} and \textit{no update}; and provided results in the supplementary material. While comparing our method over PPIDSG using different numbers of clients, we found our method better, confirming its reliability and practical utility. We have shown the effect of increasing clients for both methods using CIFAR10 and FMNIST datasets in Fig.~\ref{fig:client_compare_CIFAR-10}. \\

\noindent
\textbf{Comparison of train and test time:} Our proposed method (SBPU) takes almost the same time as PPIDSG during the training and testing phase and offers significant performance improvement. We provide a comparative analysis of the time taken by both methods on different datasets in Table~\ref{tab:training_times}, which supports its practical utility.

% During training, our method incurs minimal computation overhead over PPIDSG (refer to Table~\ref{tab:training_times}) due to the requirement of global models from the previous two FL rounds to compute bidirectional updates. Specifically, we only require additional time complexity of $O(K.L)$  ($K$=~\#Clients, $L$ =~\#Layers) for diverse models. However, there will be no computation overhead during test interference over PPIDSG as we implemented our method under the same architecture settings, which supports its practical utility.

\section{Conclusion and Future Work}  \label{sec:conclusion}
Our work makes a significant contribution to generate diverse global models for clients to improve the generalization and robustness against different privacy attacks. Our method follows the stochastic bidirectional update mechanism, which offers systematic perturbations to the global model in the parameter space of the model to generate diverse updates for clients. We validated the significance and utility of the proposed method through extensive experimentation on four datasets and surpassed the available SOTA methods. While improving privacy leakage issues during attacks, our method does not sacrifice the performance and improves the utility-security trade-off in FL. Our method offers an opportunity for future researchers to optimize existing marginal computation overhead in SBPU and explore more sophisticated bidirectional update methods. In the future, we can validate SBPU robustness across diverse data types in security-critical areas like healthcare, social media, and surveillance.

{
    \small
    \bibliographystyle{ieeenat_fullname}
    \bibliography{ref}

\begin{thebibliography}{41}
\providecommand{\natexlab}[1]{#1}
\providecommand{\url}[1]{\texttt{#1}}
\expandafter\ifx\csname urlstyle\endcsname\relax
  \providecommand{\doi}[1]{doi: #1}\else
  \providecommand{\doi}{doi: \begingroup \urlstyle{rm}\Url}\fi

\bibitem[Abadi et~al.(2016)Abadi, Chu, Goodfellow, McMahan, Mironov, Talwar, and Zhang]{abadi2016deep}
Martin Abadi, Andy Chu, Ian Goodfellow, H~Brendan McMahan, Ilya Mironov, Kunal Talwar, and Li Zhang.
\newblock Deep learning with differential privacy.
\newblock In \emph{Proceedings of the 2016 ACM SIGSAC conference on computer and communications security}, pages 308--318, 2016.

\bibitem[Aono et~al.(2017)Aono, Hayashi, Wang, Moriai, et~al.]{aono2017privacy}
Yoshinori Aono, Takuya Hayashi, Lihua Wang, Shiho Moriai, et~al.
\newblock Privacy-preserving deep learning via additively homomorphic encryption.
\newblock \emph{IEEE transactions on information forensics and security}, 13\penalty0 (5):\penalty0 1333--1345, 2017.

\bibitem[Bonawitz et~al.(2017)Bonawitz, Ivanov, Kreuter, Marcedone, McMahan, Patel, Ramage, Segal, and Seth]{bonawitz2017practical}
Keith Bonawitz, Vladimir Ivanov, Ben Kreuter, Antonio Marcedone, H~Brendan McMahan, Sarvar Patel, Daniel Ramage, Aaron Segal, and Karn Seth.
\newblock Practical secure aggregation for privacy-preserving machine learning.
\newblock In \emph{proceedings of the 2017 ACM SIGSAC Conference on Computer and Communications Security}, pages 1175--1191, 2017.

\bibitem[Chen et~al.(2020)Chen, Chen, Zhou, and Kailkhura]{chen2020fedcluster}
Cheng Chen, Ziyi Chen, Yi Zhou, and Bhavya Kailkhura.
\newblock Fedcluster: Boosting the convergence of federated learning via cluster-cycling.
\newblock In \emph{2020 IEEE International Conference on Big Data (Big Data)}, pages 5017--5026. IEEE, 2020.

\bibitem[Chuman et~al.(2018)Chuman, Sirichotedumrong, and Kiya]{chuman2018encryption}
Tatsuya Chuman, Warit Sirichotedumrong, and Hitoshi Kiya.
\newblock Encryption-then-compression systems using grayscale-based image encryption for jpeg images.
\newblock \emph{IEEE Transactions on Information Forensics and security}, 14\penalty0 (6):\penalty0 1515--1525, 2018.

\bibitem[Deng(2012)]{deng2012mnist}
Li Deng.
\newblock The mnist database of handwritten digit images for machine learning research [best of the web].
\newblock \emph{IEEE signal processing magazine}, 29\penalty0 (6):\penalty0 141--142, 2012.

\bibitem[Fraboni et~al.(2021)Fraboni, Vidal, Kameni, and Lorenzi]{fraboni2021clustered}
Yann Fraboni, Richard Vidal, Laetitia Kameni, and Marco Lorenzi.
\newblock Clustered sampling: Low-variance and improved representativity for clients selection in federated learning.
\newblock In \emph{International Conference on Machine Learning}, pages 3407--3416. PMLR, 2021.

\bibitem[Fu et~al.(2022)Fu, Zhang, Ji, Chen, Wu, Guo, Zhou, Liu, and Wang]{fu2022label}
Chong Fu, Xuhong Zhang, Shouling Ji, Jinyin Chen, Jingzheng Wu, Shanqing Guo, Jun Zhou, Alex~X Liu, and Ting Wang.
\newblock Label inference attacks against vertical federated learning.
\newblock In \emph{31st USENIX security symposium (USENIX Security 22)}, pages 1397--1414, 2022.

\bibitem[Gao et~al.(2021)Gao, Guo, Zhang, Qiu, Wen, and Liu]{gao2021privacy}
Wei Gao, Shangwei Guo, Tianwei Zhang, Han Qiu, Yonggang Wen, and Yang Liu.
\newblock Privacy-preserving collaborative learning with automatic transformation search.
\newblock In \emph{Proceedings of the IEEE/CVF Conference on Computer Vision and Pattern Recognition}, pages 114--123, 2021.

\bibitem[Geng et~al.(2021)Geng, Mou, Li, Li, Beyan, Decker, and Rong]{geng2021towards}
Jiahui Geng, Yongli Mou, Feifei Li, Qing Li, Oya Beyan, Stefan Decker, and Chunming Rong.
\newblock Towards general deep leakage in federated learning.
\newblock \emph{arXiv preprint arXiv:2110.09074}, 2021.

\bibitem[Goodfellow et~al.(2014)Goodfellow, Pouget-Abadie, Mirza, Xu, Warde-Farley, Ozair, Courville, and Bengio]{goodfellow2014generative}
Ian Goodfellow, Jean Pouget-Abadie, Mehdi Mirza, Bing Xu, David Warde-Farley, Sherjil Ozair, Aaron Courville, and Yoshua Bengio.
\newblock Generative adversarial nets.
\newblock \emph{Advances in neural information processing systems}, 27, 2014.

\bibitem[Guo et~al.(2021)Guo, Wang, Zhou, Jiang, and Patel]{guo2021multi}
Pengfei Guo, Puyang Wang, Jinyuan Zhou, Shanshan Jiang, and Vishal~M Patel.
\newblock Multi-institutional collaborations for improving deep learning-based magnetic resonance image reconstruction using federated learning.
\newblock In \emph{Proceedings of the IEEE/CVF conference on computer vision and pattern recognition}, pages 2423--2432, 2021.

\bibitem[Hitaj et~al.(2017)Hitaj, Ateniese, and Perez-Cruz]{hitaj2017deep}
Briland Hitaj, Giuseppe Ateniese, and Fernando Perez-Cruz.
\newblock Deep models under the gan: information leakage from collaborative deep learning.
\newblock In \emph{Proceedings of the 2017 ACM SIGSAC conference on computer and communications security}, pages 603--618, 2017.

\bibitem[Huang et~al.(2020)Huang, Song, Li, and Arora]{huang2020instahide}
Yangsibo Huang, Zhao Song, Kai Li, and Sanjeev Arora.
\newblock Instahide: Instance-hiding schemes for private distributed learning.
\newblock In \emph{International conference on machine learning}, pages 4507--4518. PMLR, 2020.

\bibitem[Jiang et~al.(2022)Jiang, Wang, and Dou]{jiang2022harmofl}
Meirui Jiang, Zirui Wang, and Qi Dou.
\newblock Harmofl: Harmonizing local and global drifts in federated learning on heterogeneous medical images.
\newblock In \emph{Proceedings of the AAAI Conference on Artificial Intelligence}, pages 1087--1095, 2022.

\bibitem[Jin et~al.(2023)Jin, Yao, Han, Joe-Wong, Ravi, Avestimehr, and He]{jin2023fedml}
Weizhao Jin, Yuhang Yao, Shanshan Han, Carlee Joe-Wong, Srivatsan Ravi, Salman Avestimehr, and Chaoyang He.
\newblock Fedml-he: An efficient homomorphic-encryption-based privacy-preserving federated learning system.
\newblock \emph{arXiv preprint arXiv:2303.10837}, 2023.

\bibitem[Karimireddy et~al.(2020)Karimireddy, Kale, Mohri, Reddi, Stich, and Suresh]{karimireddy2020scaffold}
Sai~Praneeth Karimireddy, Satyen Kale, Mehryar Mohri, Sashank Reddi, Sebastian Stich, and Ananda~Theertha Suresh.
\newblock Scaffold: Stochastic controlled averaging for federated learning.
\newblock In \emph{International conference on machine learning}, pages 5132--5143. PMLR, 2020.

\bibitem[Krizhevsky et~al.(2009)Krizhevsky, Hinton, et~al.]{krizhevsky2009learning}
Alex Krizhevsky, Geoffrey Hinton, et~al.
\newblock Learning multiple layers of features from tiny images.
\newblock 2009.

\bibitem[Li et~al.(2021)Li, Zhang, Wang, Han, and Li]{li2021privacy}
Anran Li, Lan Zhang, Junhao Wang, Feng Han, and Xiang-Yang Li.
\newblock Privacy-preserving efficient federated-learning model debugging.
\newblock \emph{IEEE Transactions on Parallel and Distributed Systems}, 33\penalty0 (10):\penalty0 2291--2303, 2021.

\bibitem[Li et~al.(2020)Li, Sahu, Zaheer, Sanjabi, Talwalkar, and Smith]{li2020federated}
Tian Li, Anit~Kumar Sahu, Manzil Zaheer, Maziar Sanjabi, Ameet Talwalkar, and Virginia Smith.
\newblock Federated optimization in heterogeneous networks.
\newblock \emph{Proceedings of Machine learning and systems}, 2:\penalty0 429--450, 2020.

\bibitem[Li et~al.(2019)Li, Huang, Yang, Wang, and Zhang]{li2019convergence}
Xiang Li, Kaixuan Huang, Wenhao Yang, Shusen Wang, and Zhihua Zhang.
\newblock On the convergence of fedavg on non-iid data.
\newblock \emph{arXiv preprint arXiv:1907.02189}, 2019.

\bibitem[Liao et~al.(2023)Liao, Liu, Zheng, Yao, and Chen]{liao2023ppgencdr}
Xinting Liao, Weiming Liu, Xiaolin Zheng, Binhui Yao, and Chaochao Chen.
\newblock Ppgencdr: A stable and robust framework for privacy-preserving cross-domain recommendation.
\newblock In \emph{Proceedings of the AAAI Conference on Artificial Intelligence}, pages 4453--4461, 2023.

\bibitem[Lin et~al.(2020)Lin, Kong, Stich, and Jaggi]{lin2020ensemble}
Tao Lin, Lingjing Kong, Sebastian~U Stich, and Martin Jaggi.
\newblock Ensemble distillation for robust model fusion in federated learning.
\newblock \emph{Advances in Neural Information Processing Systems}, 33:\penalty0 2351--2363, 2020.

\bibitem[Liu et~al.(2021)Liu, Chen, Qin, Dou, and Heng]{liu2021feddg}
Quande Liu, Cheng Chen, Jing Qin, Qi Dou, and Pheng-Ann Heng.
\newblock Feddg: Federated domain generalization on medical image segmentation via episodic learning in continuous frequency space.
\newblock In \emph{Proceedings of the IEEE/CVF Conference on Computer Vision and Pattern Recognition}, pages 1013--1023, 2021.

\bibitem[Ma et~al.(2024)Ma, Yao, and Xu]{ma2024ppidsg}
Yuting Ma, Yuanzhi Yao, and Xiaohua Xu.
\newblock Ppidsg: A privacy-preserving image distribution sharing scheme with gan in federated learning.
\newblock In \emph{Proceedings of the AAAI Conference on Artificial Intelligence}, pages 14272--14280, 2024.

\bibitem[McMahan et~al.(2017)McMahan, Moore, Ramage, Hampson, and y~Arcas]{mcmahan2017communication}
Brendan McMahan, Eider Moore, Daniel Ramage, Seth Hampson, and Blaise~Aguera y Arcas.
\newblock Communication-efficient learning of deep networks from decentralized data.
\newblock In \emph{Artificial intelligence and statistics}, pages 1273--1282. PMLR, 2017.

\bibitem[Sattler et~al.(2021)Sattler, Korjakow, Rischke, and Samek]{sattler2021fedaux}
Felix Sattler, Tim Korjakow, Roman Rischke, and Wojciech Samek.
\newblock Fedaux: Leveraging unlabeled auxiliary data in federated learning.
\newblock \emph{IEEE Transactions on Neural Networks and Learning Systems}, 34\penalty0 (9):\penalty0 5531--5543, 2021.

\bibitem[Sellami and Tabbone(2022)]{sellami2022deep}
Akrem Sellami and Salvatore Tabbone.
\newblock Deep neural networks-based relevant latent representation learning for hyperspectral image classification.
\newblock \emph{Pattern Recognition}, 121:\penalty0 108224, 2022.

\bibitem[Shokri et~al.(2017)Shokri, Stronati, Song, and Shmatikov]{shokri2017membership}
Reza Shokri, Marco Stronati, Congzheng Song, and Vitaly Shmatikov.
\newblock Membership inference attacks against machine learning models.
\newblock In \emph{2017 IEEE symposium on security and privacy (SP)}, pages 3--18. IEEE, 2017.

\bibitem[Sun et~al.(2021)Sun, Li, Wang, Yang, Li, and Chen]{sun2021soteria}
Jingwei Sun, Ang Li, Binghui Wang, Huanrui Yang, Hai Li, and Yiran Chen.
\newblock Soteria: Provable defense against privacy leakage in federated learning from representation perspective.
\newblock In \emph{Proceedings of the IEEE/CVF conference on computer vision and pattern recognition}, pages 9311--9319, 2021.

\bibitem[Wei et~al.(2020)Wei, Li, Ding, Ma, Yang, Farokhi, Jin, Quek, and Poor]{wei2020federated}
Kang Wei, Jun Li, Ming Ding, Chuan Ma, Howard~H Yang, Farhad Farokhi, Shi Jin, Tony~QS Quek, and H~Vincent Poor.
\newblock Federated learning with differential privacy: Algorithms and performance analysis.
\newblock \emph{IEEE transactions on information forensics and security}, 15:\penalty0 3454--3469, 2020.

\bibitem[Wu et~al.(2021)Wu, Kang, Luo, He, and Yang]{wu2021fedcg}
Yuezhou Wu, Yan Kang, Jiahuan Luo, Yuanqin He, and Qiang Yang.
\newblock Fedcg: Leverage conditional gan for protecting privacy and maintaining competitive performance in federated learning.
\newblock \emph{arXiv preprint arXiv:2111.08211}, 2021.

\bibitem[Xiao et~al.(2017)Xiao, Rasul, and Vollgraf]{xiao2017fashion}
Han Xiao, Kashif Rasul, and Roland Vollgraf.
\newblock Fashion-mnist: a novel image dataset for benchmarking machine learning algorithms.
\newblock \emph{arXiv preprint arXiv:1708.07747}, 2017.

\bibitem[Xie et~al.(2022)Xie, Zhang, Li, Kang, Niyato, Xie, and Wu]{xie2022efficient}
Kan Xie, Zhe Zhang, Bo Li, Jiawen Kang, Dusit Niyato, Shengli Xie, and Yi Wu.
\newblock Efficient federated learning with spike neural networks for traffic sign recognition.
\newblock \emph{IEEE Transactions on Vehicular Technology}, 71\penalty0 (9):\penalty0 9980--9992, 2022.

\bibitem[Yin et~al.(2021)Yin, Mallya, Vahdat, Alvarez, Kautz, and Molchanov]{yin2021see}
Hongxu Yin, Arun Mallya, Arash Vahdat, Jose~M Alvarez, Jan Kautz, and Pavlo Molchanov.
\newblock See through gradients: Image batch recovery via gradinversion.
\newblock In \emph{Proceedings of the IEEE/CVF Conference on Computer Vision and Pattern Recognition}, pages 16337--16346, 2021.

\bibitem[Yu et~al.(2023)Yu, Liu, Wu, Yu, Yu, and Zhang]{yu2023untargeted}
Yang Yu, Qi Liu, Likang Wu, Runlong Yu, Sanshi~Lei Yu, and Zaixi Zhang.
\newblock Untargeted attack against federated recommendation systems via poisonous item embeddings and the defense.
\newblock In \emph{Proceedings of the AAAI Conference on Artificial Intelligence}, pages 4854--4863, 2023.

\bibitem[Yuval(2011)]{yuval2011reading}
Netzer Yuval.
\newblock Reading digits in natural images with unsupervised feature learning.
\newblock In \emph{Proceedings of the NIPS Workshop on Deep Learning and Unsupervised Feature Learning}, 2011.

\bibitem[Zhao et~al.(2020)Zhao, Mopuri, and Bilen]{zhao2020idlg}
Bo Zhao, Konda~Reddy Mopuri, and Hakan Bilen.
\newblock idlg: Improved deep leakage from gradients.
\newblock \emph{arXiv preprint arXiv:2001.02610}, 2020.

\bibitem[Zhu et~al.(2023)Zhu, Yao, and Blaschko]{zhu2023surrogate}
Junyi Zhu, Ruicong Yao, and Matthew~B Blaschko.
\newblock Surrogate model extension (sme): A fast and accurate weight update attack on federated learning.
\newblock \emph{arXiv preprint arXiv:2306.00127}, 2023.

\bibitem[Zhu et~al.(2019)Zhu, Liu, and Han]{zhu2019deep}
Ligeng Zhu, Zhijian Liu, and Song Han.
\newblock Deep leakage from gradients.
\newblock \emph{Advances in neural information processing systems}, 32, 2019.

\bibitem[Zhu et~al.(2021)Zhu, Hong, and Zhou]{zhu2021data}
Zhuangdi Zhu, Junyuan Hong, and Jiayu Zhou.
\newblock Data-free knowledge distillation for heterogeneous federated learning.
\newblock In \emph{International conference on machine learning}, pages 12878--12889. PMLR, 2021.

\end{thebibliography}
}

\end{document}